\documentclass{article}
\usepackage{geometry} 
\usepackage{graphicx}
\graphicspath{ {../figures/} }
\usepackage{caption}
\usepackage{subcaption}
\usepackage[english]{babel}
\usepackage[T1]{fontenc}
\usepackage{ragged2e}
\usepackage{chemformula}
\usepackage{soul}
\usepackage{xcolor}
\usepackage{array}
\usepackage[numbers,sort&compress]{natbib}

\usepackage{hyperref}
\usepackage{changepage} 
\definecolor{lightgrey}{rgb}{0.95,0.95,0.95}
\definecolor{kgrey}{rgb}{0.6,0.6,0.6}
\definecolor{lgreen}{rgb}{0.88, 1, 0.88} 
\usepackage{colortbl} 
\usepackage{multirow} 
\usepackage{multicol}

\usepackage{tabularx}
\newcolumntype{C}{>{\centering\arraybackslash}X}  
\newcolumntype{L}{>{\raggedright\arraybackslash}X} 
\newcolumntype{M}[1]{>{\centering\arraybackslash}m{#1}}

\usepackage{makecell}
\definecolor{goodgreen}{rgb}{0.88, 1, 0.88} 
\definecolor{badred}{rgb}{1, 0.88, 0.88}    
\definecolor{cream}{RGB}{222,217,201}

\begin{document}

{\centering
\null \vspace{3.5 cm}
\noindent \textbf{\huge Towards a Science Exocortex}
\par \null \vspace{1.5 cm}
\noindent \textbf{\Large }
\par \null \vspace{1.5 cm}
\noindent {\Large Kevin G. Yager$^{\ast}$}
\par \null \vspace{0.5 cm}
\noindent {\large Center for Functional Nanomaterials, Brookhaven National Laboratory, Upton, New York 11973, United States}
\par\null \vspace{0.5 cm}
\par\null\par
}

\par


\begin{abstract}
\noindent 
\begin{adjustwidth}{1cm}{1cm} 
Artificial intelligence (AI) methods are poised to revolutionize intellectual work, with generative AI enabling automation of text analysis, text generation, and simple decision making or reasoning. The impact to science is only just beginning, but the opportunity is significant since scientific research relies fundamentally on extended chains of cognitive work. Here, we review the state of the art in agentic AI systems, and discuss how these methods could be extended to have even greater impact on science. We propose the development of an exocortex, a synthetic extension of a person's cognition. A science exocortex could be designed as a swarm of AI agents, with each agent individually streamlining specific researcher tasks, and whose inter-communication leads to emergent behavior that greatly extend the researcher's cognition and volition.
\end{adjustwidth}
\end{abstract}

\section{Introduction}
Artificial intelligence and machine-learning (AI/ML) methods are having growing impact across a wide range of field, including the physical sciences.\cite{Qiu2016, ai4science2023impact, Wang2023scienceAgeAI, royalsociety2024AI, D3DD00113J, FosterReview} 
Generative foundation models, in particular, are displacing a swath of other methods. 
Foundation models involve extensive training of deep neural networks on enormous datasets in a task-agnostic manner.\cite{NEURIPS2020_1457c0d6, foundation2021} 
Generative methods (genAI), often employing the transformer architecture,\cite{vaswani2017attention} seek to create novel outputs that conform to the statistical structure of training data,\cite{9903869, gozalobrizuela2023chatgpt} enabling (e.g.) image synthesis\cite{DALLE2, rombach2021highresolution, Oppenlaender2022} or text generation.\cite{Radford2018} 
Large language models (LLMs) are generative models trained on text completion, but which can be adapted to a variety of tasks, including text classification, sentiment analysis, code or document generation, or  interactive chatbots that respond to users in natural language.\cite{NEURIPS2020_1457c0d6, yang2023harnessing,liu2024understanding,minaee2024large} 
The performance of LLMs increases with the scale of the training data, network size, and training time.\cite{hestness2017deep, Henighan2020-10-27, hoffmann2022training} 
There is growing evidence that LLMs do not merely reproduce surface statistics, but learn a meaningful world model;\cite{li2023emergent, akyürek2023learning, kosinski2023theory, Webb2023, gurnee2024language, vafa2024evaluating} one correspondingly observes sudden leaps in capabilities during training, suggesting the emergent learning of generalized concepts.\cite{Ganguli_2022, wei2022emergent, nanda2023progress, bubeck2023sparks, Webb2023} 
LLMs can be tailored via reinforcement learning using human feedback (RLHF),\cite{ziegler2020finetuning, ouyang2022training, lambert2022illustrating, lee2023rlaif} so that particular behaviors (e.g. helpful and truthful) are emphasized during generation. 
Generation quality can be improved by connecting to a corpus of trusted documents, which allows production of replies that are sourced and grounded (so-called retrieval augmented generation, RAG).\cite{2020RAG,Yager2024chatbot,gao2024retrievalaugmented,yu2024evaluation}

\begin{figure*}
\centering
  \includegraphics[width=15.0cm]{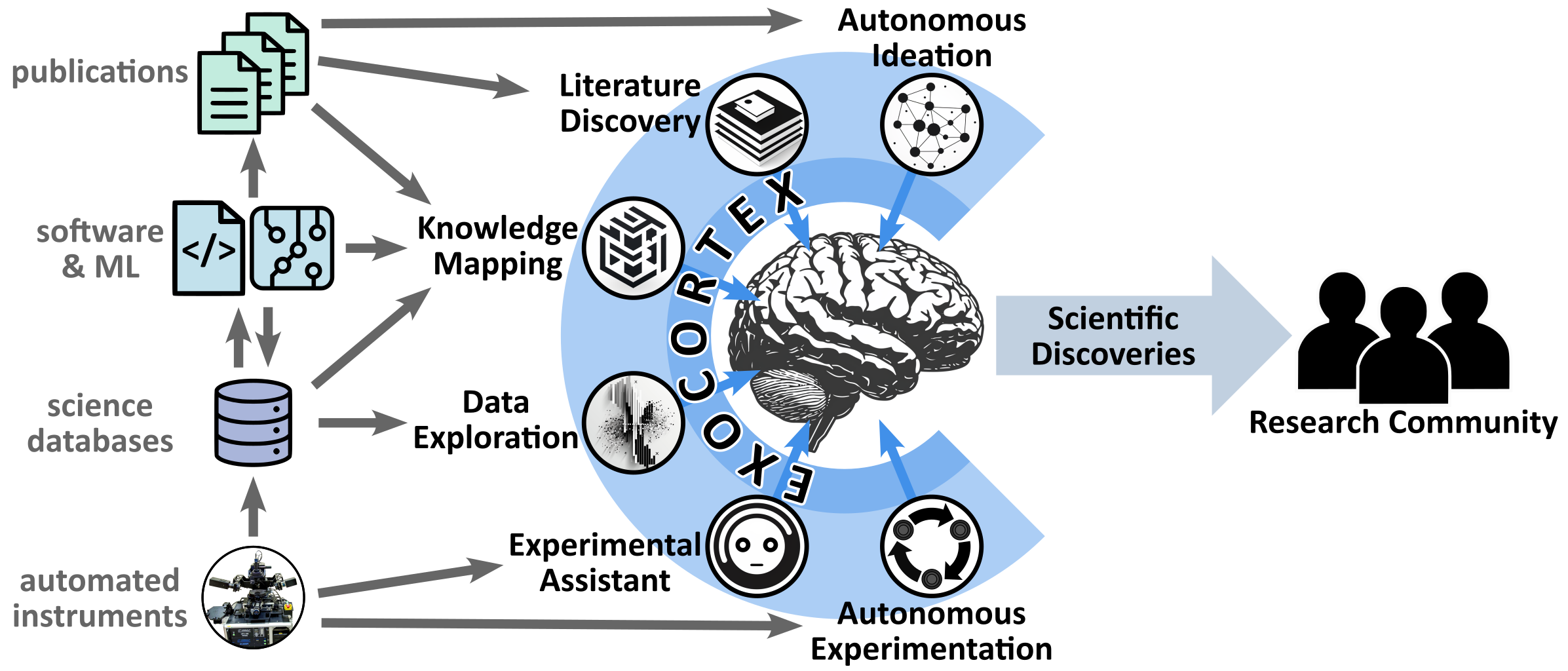}
  \caption{An exocortex seeks to augment human intelligence by connecting computation systems to a person. A science exocortex could be implemented as a swarm of specialized AI agents, operating on behalf of the human researcher, including agents for controlling experimental systems, for exploring data and synthesizing it into knowledge, and for exploring literature and ideation. The AI agents would connect to science components (instruments, databases, software, etc.) and streamline access. Crucially, the AI agents communicate with one another, working on tasks on behalf of the user and only surfacing the most important decisions and outputs for human consideration. If successful, such a system would allow researchers to handle the enormity of modern scientific knowledge, and accelerate discovery and dissemination of new science.
}
  \label{fgr:exocortex}
\end{figure*}

\begin{figure*}
\centering
  \includegraphics[width=8.0cm]{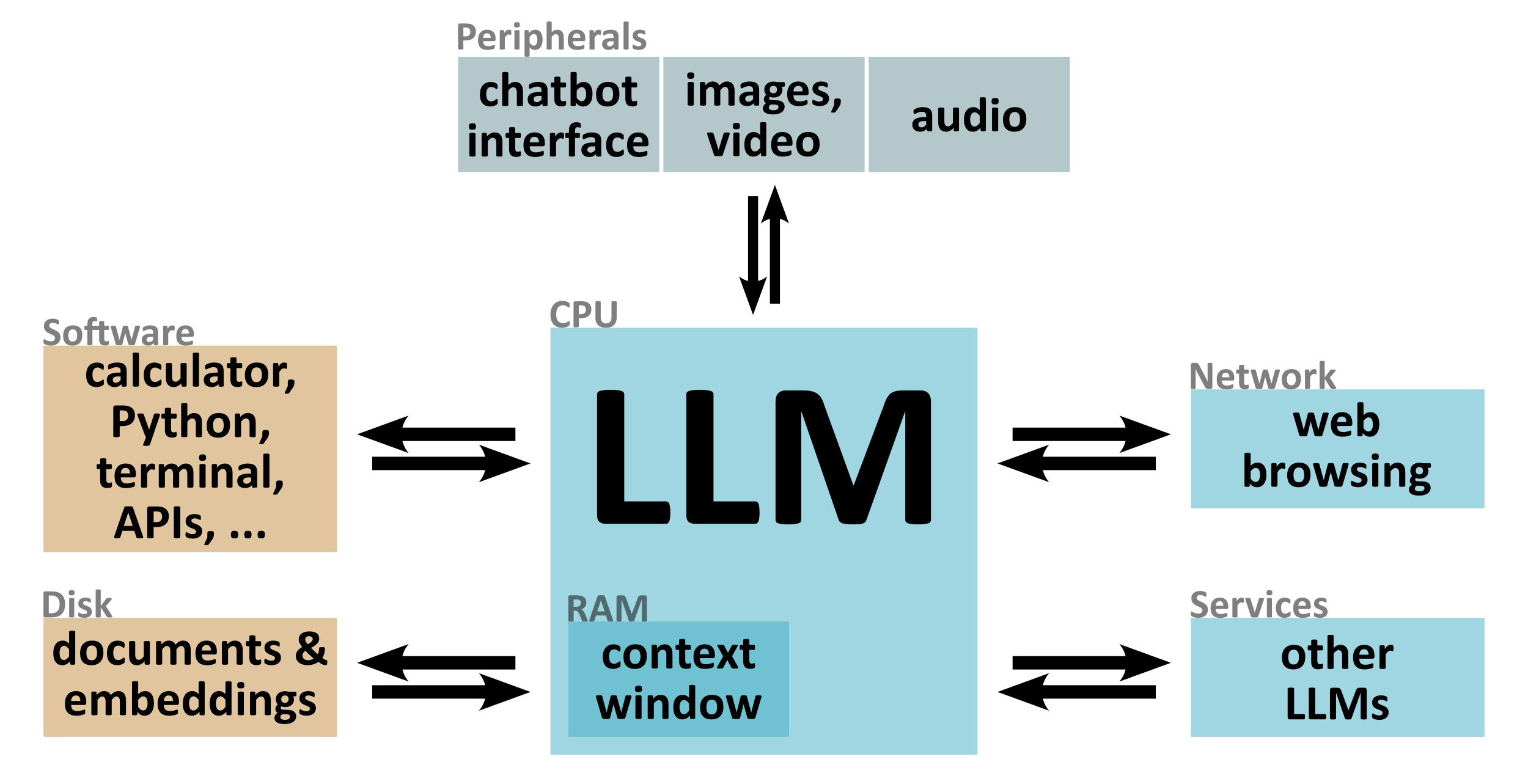}
  \caption{Diagram of a large language model (LLM) acting as a kernel (image based on social media post by Andrej Karpathy). While LLMs perform text generation, Karpathy has proposed to view them instead as kernels---orchestration agents---of a new kind of operating system.\cite{karpathy2023llmos, karpathy2023llmos2} In this paradigm, the LLM is responsible for accessing resources (e.g. documents) or triggering actions (calculations, web browsing, etc.), and feeding results to a desired interface (e.g. chatbot dialog). The ability of LLMs to perform (rudimentary) decision-making can thus be exploited to coordinate more complex activity in response to relatively vague commands (which may come from a human or another LLM system).
}
  \label{fgr:kernel}
\end{figure*}

While LLMs are often viewed purely as text-generators (e.g. for chat interactions), they have transformative potential owing to their ability to generate decisions and plans. For instance, LLMs can trigger software tools by providing them access to application programming interfaces (APIs).\cite{yao2023react, schick2023toolformer, gao2023pal, liang2023taskmatrixai, shen2023hugginggpt, cai2023large, peng2023check, xu2023rewoo, hsieh2023tool}
Generations can be improved by inducing self-critique of output quality,\cite{shinn2023reflexion, lightman2023lets, McAleese2024LLMBugs} or creating chains of thought through iterative self-prompting.\cite{xu2023reprompting, yao2023tree, xu2024faithful} 
These systems can be turned into task-oriented autonomous agents by allowing them to iteratively propose and execute solutions.\cite{shen2023hugginggpt, wang2023voyager, li2023camel, Boiko2023Nature, yang2024selfgoal}

The impressive capabilities of LLMs presage a paradigm shift in the way intellectual work is performed, as they empower humans to delegate many tasks to the LLM and instead focus on the highest-level deliberation and planning. 
However, there remain many outstanding questions about what system architecture and human-computer interactions (HCI) will best leverage these capabilities. 
Adaptation of these methods to scientific domains requires even deeper consideration, as science and engineering tasks are extremely technical and require high reliability and sourcing for both information and arguments.

Here, we explore the concept of an exocortex---an artificial extension to the human brain that provides additional cognitive capabilities. While future implementations of this concept might employ brain-computer interfaces (BCIs),\cite{6893912} we argue that progress can be made by leveraging existing HCI methods to connect the human to a swarm of inter-communicating AI agents. If the individual agents are sufficiently capable, and their interactions sufficiently coherent, then the emergent activity could feel, to the human operator, as an empowering expansion to their mental capabilities. 

We focus in particular on the concept of a \textit{science} exocortex---meant to expand a researcher's intelligence and scientific reasoning---and propose some concrete architectural ideas. We propose an implementation (Figure~\ref{fgr:exocortex}) using a swarm of AI agents that operate on behalf of the human user, and which---importantly---communicate with one another and thereby reserve human interaction only for high-value ideas and important decisions. 
We define specific categories of required agents, including some focused on orchestrating experiments, others on data and software, and others on scientific literature. 
Although highly speculative, we hope the ideas presented herein stimulate further research on AI agents optimized for science, and their integration into systems that empower human researchers.

\section{Discussion}

LLMs natively output streams of tokens, and are by default used to generate text for humans to read, as in the canonical use as chatbots. However, a narrow interpretation of LLMs would miss their most significant capability: their outputs can be used as decisions, allowing one to automate (simple) cognitive tasks. 
Karpathy provides a provocative vision for the future of LLMs, wherein they act as kernels (orchestration agents) of a diverse set of capabilities (Figure~\ref{fgr:kernel}).\cite{karpathy2023llmos, karpathy2023llmos2} In a conventional operating system (OS), the kernel is a privileged software process that manages resource distribution and inter-process communication, allowing the end-user to access software systems, files on disk, network resources, and other services. By analogy, one can imagine a sort of AI OS, where the orchestration abilities of the LLM are leveraged to intelligently trigger the appropriate tool (via APIs,\cite{yao2023react, schick2023toolformer, gao2023pal, liang2023taskmatrixai, shen2023hugginggpt, cai2023large, peng2023check, xu2023rewoo, hsieh2023tool} code execution, etc.), retrieve relevant content (via RAG\cite{2020RAG,Yager2024chatbot,gao2024retrievalaugmented,yu2024evaluation}, web browsing, etc.), and reformulate it into a form suitable for human consumption (text, images, audio, etc.). The crucial insight is that the LLM enables orchestration of tasks and resources, and aggregation of data sources, in a much more abstracted and high-level manner than is traditionally thought of as possible for software systems.

The exocortex concept takes this idea seriously, and expands upon it to propose that a swarm of agents could handle complex tasks. Each agent would operate in the manner depicted in Figure~\ref{fgr:kernel}, optimized for a particular task (by tailoring the available tools/documents, the prompting and scaffolding that dictate its input/output behavior, etc.). The interaction of agents, each acting as a sort of primitive cognitive module, could then lead to emergent capabilities in the whole.

Achieving this vision will be difficult, requiring solving a cascade of research challenges. Research is required to determine how best to exploit LLMs to generate agentic modules that can perform tasks autonomously (over short timescales) by iterating on a problem. Specialization of these agents to scientific problems will require additional consideration. The software infrastructure to have agents run over longer time periods, and inter-communicate productively, will need to be developed. The correct inter-agent organizational and communication structure will need to be identified. And, finally, the appropriate interface between the ecosystem of AI agents, and the human operator, will need to be developed. 
Below we provide initial thoughts on these various challenges.

\subsection{AI Agents}

Research into LLM-based AI agents is ongoing, with several prototypes having been demonstrated.\cite{shen2023hugginggpt, wang2023voyager, li2023camel, Boiko2023Nature, xi2023rise, Wang_2024, ramos2024reviewlargelanguagemodels, jin2024llmsllmbasedagentssoftware, kapoor2024aiagentsmatter} 
Although the optimal architecture remains an open question, current demos typically add several elements to the base LLM, such as: providing the LLM with access to various tools (software, web browsing, etc.), the ability to store information about its ongoing work (i.e. memory\cite{zhong2023memorybank, wang2023augmenting, das2024larimar, Li2024Banishing}), some kind of loop to iterate on problems (inner monologue\cite{zelikman2024quietstar} or chain-of-thought\cite{lightman2023lets, xu2023reprompting, yao2023tree, xu2024faithful} or internal graph search\cite{bounsi2024transformers}), and prompting suggesting breaking a problem into steps, and then working progressively on each step. 
Further improvements are possible using an architecture where task plans and status are captured in an explicit tree structure, which provides a flexible way to organize complex hierarchies.\cite{yang2024selfgoal} Tree structures can be efficiently searched (e.g. Monte Carlo tree search) for reasoning and planning, yielding improvements in many tasks including math.\cite{luo2024improve, chen2024alphamath, zhang2024restmcts, zhang2024accessing, koh2024tree}

Additional research will be required to adapt agentic LLM approaches to scientific problems. Straightforward improvements would arise from training or fine-tuning LLMs on scientific documents, to ensure understanding of the relevant topics. Fine-tuning on math examples can elicit latent mathematical abilities.\cite{li2024common7blanguagemodels, zhang2024mathtuning} Document retrieval can also easily improve LLM performance on scientific tasks.\cite{Yager2024chatbot} 
Additional LLM specializations for science should also be considered. Golkar et al. proposed xVal, a specialized token encoding for numbers (scaling a dedicated embedding vector) which improves LLM handling of numerical tasks.\cite{golkar2023xval} McLeish et al. used special positional embeddings (relative to start of number) and demonstrated vastly improved performance and generalization on simple arithmetic (addition and multiplication) tasks.\cite{mcleish2024transformers} 
Xu et al. integrated symbolic expressions and logic rules into a chain-of-thought prompting strategy, demonstrating improved reasoning on logical tasks since the LLM was invoking formal logic and symbol manipulation during the solution.\cite{xu2024faithful} 
Trinh et al. combined a language model with a symbolic solver to handle geometry theorems.\cite{Trinh2024} 
Vashishtha et al. improved causal reasoning by providing axiomatic training examples.\cite{vashishtha2024teachingtransformerscausalreasoning}
These kinds of approaches appear promising, suggesting that LLMs with slight adaptations could yield vastly improved reasoning for science and engineering tasks.

An advantage of the exocortex architecture is that it can easily integrate more advanced AI agents as they are developed by others. In other words, we propose to separate the design/function of agents from their inter-communication, so that new agents can be added to the exocortex easily (by simply building a wrapper that supports the expected messaging between agents). The goal is to be able to leverage the growing ecosystem of AI modules being developed for science, including for simulating complex systems,\cite{emami2024syscaps} optimizing differential equations,\cite{kantamneni2024optpdediscoveringnovelintegrable} fluid dynamics,\cite{Kumar2023MYCRUNCHGPT} material discovery,\cite{jia2024llmatdesignautonomousmaterialsdiscovery} crystallography,\cite{Maffettone2021}, 2D materials,\cite{Tritsaris2021Moire} chemistry tools,\cite{bran2023chemcrow, ramos2024reviewlargelanguagemodels} protein representation\cite{xu2023protst} and design,\cite{liu2023textguided} and pathology images.\cite{Lu2024, Royer2024}

\subsubsection{Autonomous Experimentation}

Autonomous experimentation (AE) is an emerging paradigm for accelerating scientific discovery, leveraging AI/ML to automate the entire experimental loop, notably the decision-making step.\cite{Yager2018BD, Yager2023AutonomousChapter} AE aims not to merely quantitatively accelerate, but also to qualitatively improve experiment execution by having an algorithm adaptively select optimal experiments. It seeks not to replace the human researcher, but to liberate them to operate at a higher level of abstraction where they can focus on scientific meaning instead of micro-managing experimental details.\cite{Yager2023AutonomousChapter, AutonomySurveyOpinions} 
Progress in AE has grown rapidly over the last few years, transitioning from proof-of-principle to true discovery of new science.\cite{King2004, Kusne2014, Nikolaev2014, Xue2016, Ren2018, stein_progress, noack_kriging-based_2019, noack_advances_2020, noack_autonomous_2020, noack_gaussian_2021, Kalinin2021, STACH20212702, Chen2022, Zhao2022, Doerk2022AE, Bae2023, Yager_2023, Volk2023, Szymanski2023} 
The AI control module may exploit reinforcement learning,\cite{Alexander2021, Chen2022} though a highly popular approach is to exploit Bayesian methods\cite{rainforth2023modern} (such as a Gaussian process,\cite{noack_gaussian_2021, Noack2023GaussianChapter} GP) since this provides rigorous modeling of a data surrogate and associated uncertainty. 
AE methods are increasing in sophistication, including demonstration of multi-modal autonomous experiments integrating multiple measurement systems.\cite{maffettone2023selfdriving} 

Instead of treating the AE system as an autonomous loop initiated and monitored by the human researcher, one can envision it as a module in the exocortex, which can be activated and monitored by other AI agents. Enabling this capability would require relatively little change to existing AE architectures. Primarily, one would need to define a simple software API or natural-language interface for AE parameters and actions. 
Doing so would increase the power of AE systems, as they could more easily integrate physics-informed priors arising from literature or preexisting datasets.

\subsubsection{Experimental Assistant}

A highly consequential type of AI agent for science is one that negotiates control of some experimental tool on behalf of the researcher. Building such an agent requires the experimental system to already be highly automated, such that the agent can trigger operations (synthesis, measurement, etc.) and retrieve generated data. However, as more and more platforms naturally shift towards higher levels of automation, the prospects for AI control improve. 
Many high-end measurement tools provide highly software-driven interfaces, including electron microscopes,\cite{Kalinin2021, STACH20212702} scanning probe instruments,\cite{Zahl2010GXSM, Kalinin2021, Liu2022} and synchrotron\cite{Hill_2020, Yager_2023} or free electron laser (FEL)\cite{RevModPhys.88.015007, BOSTEDT2009108, Allaria2012} beamlines.
Recent work has also demonstrated automated workflows\cite{Chan2010, Li2012, yan_solar, Granda2018, Volk2023}  or modular platforms\cite{D3DD00142C, Shimizu2020, Abolhasani2023} for lab experiments. The rapidly advancing capabilities of AI robotic control\cite{rt22023arxiv, chi2023diffusionpolicy, avetisyan2024scenescript, radosavovic2024humanoid, ahn2024autort, Aldarondo2024, fu2024humanplus} suggest that broader ranges of manual laboratory tasks will soon be amenable to automation.

The next step is thus to design AI agents that can access the capabilities of these automated systems. 
Preliminary work has already demonstrated the viability of LLM-based agents for controlling scientific instruments.\cite{prince2023opportunities,potemkin2023virtual,bran2023chemcrow,liu2024synergizing} 
AI experimental assistants can allow the human to phrase commands in natural-language, whereupon the LLM can convert this into action through API calls or code generation.\cite{D3DD00113J} 
The assistant can help to integrate experimental and data analysis steps, making it easier to see the consequences of measurements and to iterate more quickly on the problem being studied. 
Experimental assistants can also act as tutors, e.g. generating initial control code for a user unfamiliar with a particular instrument. The approach is flexible, and can easily be adapted to changing instrument conditions by updating documentation that is added to the LLM's context during operation.

\subsubsection{Data Exploration}

Scientific discovery involves collecting, processing, and analyzing datasets of many types. 
In answering a scientific problem, researchers will integrate a wide variety of data sources, including lab notes, instrument outputs (images, spectra, etc.), simulation results, and a succession of derivative data products created through analysis. 
Tracking, organizing, and visualizing these datasets is extremely challenging. An acute challenge for a modern researcher is the interdisciplinary nature of many frontier topics, which correspondingly means dealing with a heterogeneity of datasets coming from different sources, and following different conventions for formatting and meta-data. 

AI assistants could play an important role in alleviating this burden, by automating many routine tasks in data triage and reformatting, and by automatically triggering the required pipeline for automated data processing. 
The heterogeneity of data can possibly be handled using foundation models.\cite{foundation2021} Whereas in the past application of machine-learning to science required training bespoke models on carefully-labelled datasets specific that science topic, foundation models trained on vast quantities of unlabelled data should be able to learn generic representations that are useful. 
For instance, it was found that the Contrastive Language-Image Pre-training (CLIP) model\cite{radford2021learning} trained on generic Internet image data could be used to assess similarity for scanning electron microscopy and x-ray scattering datasets, without any retraining or fine-tuning.\cite{Yager2024chatbot}

Many scientific datasets are images, or can be converted into images. Thus a powerful approach for AI data assistants is to exploit multi-modal language/vision models.\cite{lu2019vilbert, radford2021learning, yang2023dawn, Royer2024, carolan2024review} 
Indeed, humans generally consume data as images, in the form of graphs and plots; this visual formulation of the data has the advantage of being ready-to-deploy, human-readable, and already well-represented in existing training sources (publications). 
Exploiting multi-modal models as data assistants is still in its infancy, but early systems\cite{gao2023ophglm, NEURIPS2023_5abcdf8e, Wang2024.04.11.588958, chen2024llaga, song2024mmaccopilot, Lu2024} show promise.

\subsubsection{Knowledge Mapping}

A grand challenge in data science is to integrate data from disparate sources into a single model. 
Human scientists excel at this task, as they integrate insights provided from experimental data, calculations, literature they recall, and intuitions informed by years of scientific practice. When thinking about or discussing a complex topic, human scientists will naturally jump between different levels of abstraction and thus different scientific models. This combination of models (some highly quantitative, others heuristic) allows human scientists to compensate for the deficiencies in one model/reasoning by leveraging another. Approximating this efficient behavior in a synthetic knowledge system is challenging. 
AI analogs of human synthesis would make knowledge integration explicit and documented, and provide integrated models for other software systems to leverage.

A core challenge is to align disparate observations of the same physical signal; that is, to account for the unknown disparity between models arising from systematic errors, different underlying assumptions, mismatched definitions, etc. Consider a simple signal for a material system (e.g. crystalline grain size) as a function of a physical parameter (e.g. temperature). Despite the independent/dependent variables being well-defined within one model (observation), matching between models may not be trivial. For instance, a physical measurement might use absolute real-world temperature (in Kelvin units) while an associated coarse-grained simulation might rely on a unitless abstracted temperature variable. (Obviously the two quantities are closely related; but the mapping function between them is typically not known.) The measurement of grain size by different techniques may not match owing to different definitions (e.g. volumetric vs. aerial averaging).\cite{Majewski_2016} 
Thus, it can be challenging to merge datasets relevant to the same physical problem, even when they are individually trustworthy and robust. The problem becomes harder still as input data sources become more heterogeneous and ill-defined (heuristic classifications, text descriptions, scientist intuitions, etc.).

The simplest approach to this problem might be to exploit contrastive learning. For instance, the CLIP\cite{radford2021learning} model uses two encoder pathways: one for text and one for images. The method also computes a similarity matrix between the two latent spaces (cosine similarity for text/image pairs), where part of the training loss seeks to maximize the diagonal and minimize the off-diagonal elements. In this way, the text and image latent spaces align, allowing cross-modal learning and applications. 
In principle, a similar approach could be used for scientific data. Datasets and their associated text descriptions could be used as training pairs, or different observations of the same physical phenomenon (e.g. experimental measurements and corresponding simulations) could be combined if some pairwise associations were manually identified. 
Recent work increasing the number of modalities appears promising.\cite{4m21proceeding, bachmann20244m21} 
Nevertheless, it may be challenging to scale this approach by itself, to handle the heterogeneity, complexity, and sparsity of realistic laboratory datasets. 

A more sophisticated approach to this problem is to train multi-modal foundation models on scientific datasets.\cite{FosterReview} The Polymathic AI effort is proposing to train AI models for science on a breadth of data,\cite{PolymathicAI2024} which can then be specialized for any particular application by exploiting the latent representations or via problem-specific fine-tuning. Cranmer argues that doing otherwise (e.g. training an ML model for science using random initialization) is inefficient as it ignores the wealth of well-understood scientific priors.\cite{Cranmer2024} Initial results for this approach are promising,\cite{lanusse2023astroclip, mccabe2023multiple} with (e.g.) multi-physics pretraining on system dynamics improving subsequent predictions on new systems. The approach involves projecting the fields for different kinds of physical systems into a shared embedding space. The central generative model (based on transformers) thus learns meaningful physics, while the dataset-specific embedding/normalization schemes capture the differences between the physical systems. 

A closely-related approach would be to train multi-modal foundation models on science data, so that these models could be queried to explore trends in the data. Recent work\cite{treutlein2024connecting} has shown that an LLM trained on $(x,y)$ pairs can articulate the function $f(x)$ that underlies the transformation (can define it in code, can invert it, etc.). If this result generalizes, it implies that LLMs trained on raw science data could coherently describe the data, make predictions based on the underlying functions, and so on.

\begin{figure*}
\centering
  \includegraphics[width=10.0cm]{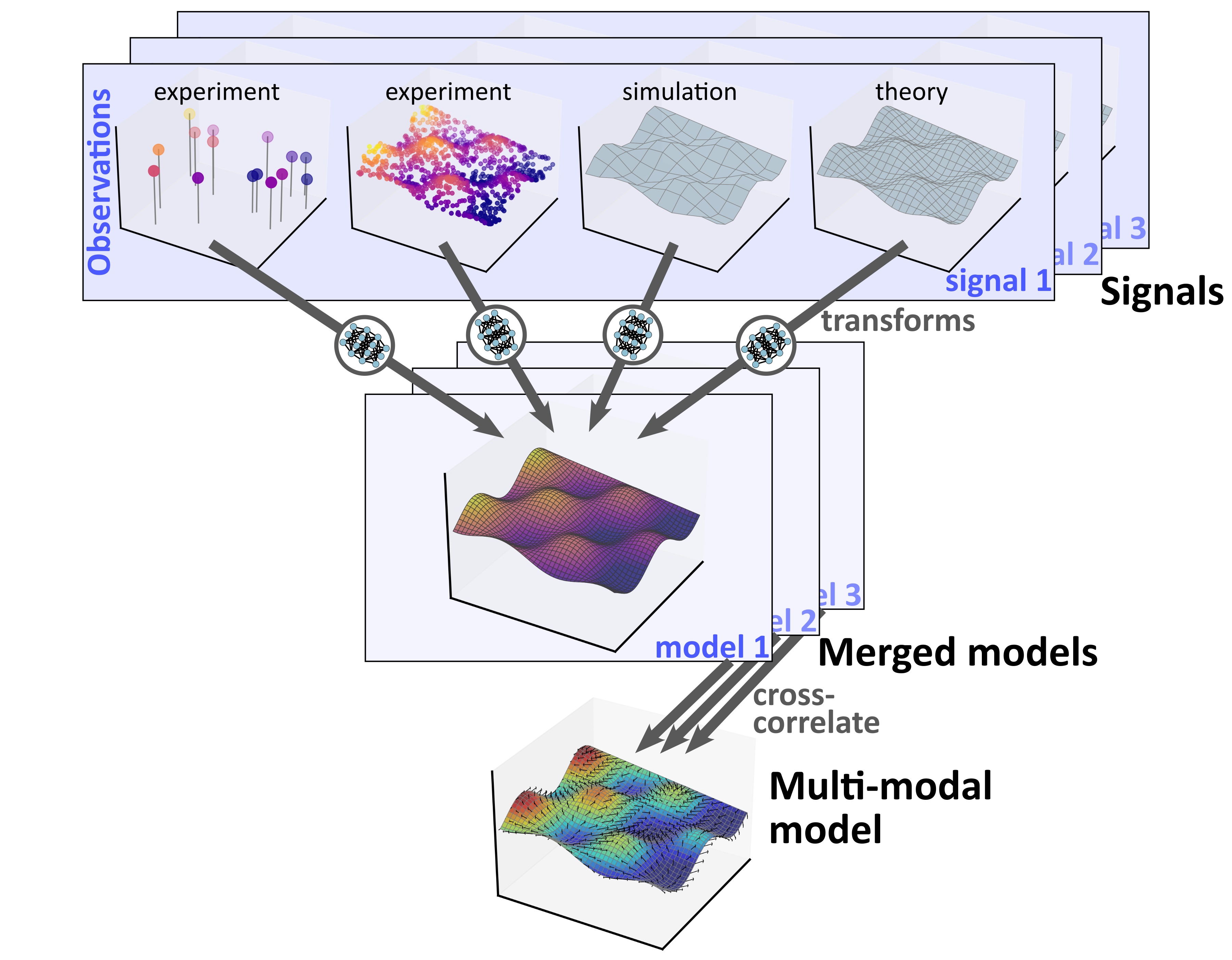}
  \caption{Knowledge mapping is an attempt to align and aggregate a variety of data sources about a particular scientific problem into a single model. One architecture for accomplishing this is shown. Available data is organized into signals of interest (such as physical measurables, material properties, or functional metrics). One typically has a variety of estimates or observations for a given signal, arising from different experiments, calculations, or theories. In principle these observations already map into a common space; in practice there are complex and often unknown disparities between the observations, owing to measurement errors, disparate definitions, or different assumptions. Thus, some non-linear transformation (e.g. accomplished using neural networks) is required to combine them into a single predictive model. Models for distinct signals can be cross-correlated to identify inter-relations; this can effectively combine the models into a single multi-modal model.
}
  \label{fgr:knowledge}
\end{figure*}

A different way to formulate this task (Figure~\ref{fgr:knowledge}) is to focus on integrated modeling of all the signals defined in physical parameter spaces for a given problem (e.g. class of materials). For a given signal, a variety of different observations might be available (from experiments, simulation, theory, etc.), with tradeoffs between signals (in terms of sampling density, error bars, validity in different parts of parameter space, etc.). 
Signals could be combined into a merged model by learning a non-linear transformation that maps them into a common space and maximizes their overlap (using, e.g., variants of the methods described above). The combined datasets could be interpolated using a Gaussian process or other nonparametric method,\cite{noack_gaussian_2021, Noack2023GaussianChapter} leveraging physics-informed constraints to further improve the model (e.g. via tailored kernel design\cite{Noack2022kernel, Bae2023}) and acceleration methods to reduce computation cost.\cite{Noack2021HGDL, Noack2022sparsity} 
The set of models could also be cross-correlated, to identify connections and scientific trends between signals (or establishing lack thereof). GP modeling of correlated signals would also allow interpolation of signals into parts of spaces where they were not explicitly measured (effectively using a correlated measurement as a sort of proxy signal). Conceptually, the set of signals (and covariance matrix between them) represents a final rich multi-modal model of full system behavior. 
This unified model could be used for predictions, searching for trends and novel physics, or as a guide for future discovery (identifying under-sampled regions, suggesting high-performing materials, etc.).

An even more speculative approach would be to attempt to adapt methods of generative world synthesis to scientific data. There has been enormous progress in generative synthesis of images (2D data),\cite{DALLE2, rombach2021highresolution, Oppenlaender2022} objects (3D),\cite{poole2022dreamfusion, wang2023prolificdreamer, poole2022dreamfusion, lin2023magic3d, metzer2022latentnerf, chen2023fantasia3d, tsalicoglou2023textmesh, liu2023zero1to3, qian2023magic123, haque2023instructnerf2nerf, gao2024cat3d} and video (3D).\cite{singer2022makeavideo, ho2022imagen, blattmann2023align, gupta2023photorealistic, kondratyuk2024videopoet, videoworldsimulators2024} Neural radiance fields\cite{Mildenhall2020Nerf} and Gaussian splatting\cite{kerbl20233d} have emerged as efficient methods for reconstructing and representing 3D scenes (where input images act as projective constraints). These methods have been extended to capture\cite{wu20234d, li2024spacetime, ren2024l4gm} or synthesize\cite{shao2023control4d, peng2024papr, yu20244real, wang2024vidu4d, pang2024ash} changes over time (4D). 
These methods are efficient,\cite{duckworth2024smerf, peng2024rtgslam} scalable,\cite{lin2024vastgaussian} and amenable to in/out-painting.\cite{weber2023nerfiller, seo2024genwarp} In addition to obvious applications in content generation, these methods are seeing adoption for autonomous driving\cite{WayvePRISM1} and robotics.\cite{3DAwareManip, EmbodiedGS, li2024objectaware, xue2024neural} 
The trajectory of these development suggests neural synthesis of virtual world,\cite{videoworldsimulators2024} wherein immersive 3D environments are generated and animated/evolved, using real-world reconstructions and/or user text commands as inputs. 
We suggest that this approach could be applied, in higher-dimensional spaces, to scientific datasets. The partial measurements made in scientific experiments can act as constraints (conceptually a projective view of the full higher-dimensional space), where the objective is to reconstruct a consolidated model consistent with all the data (i.e. merge datasets and modalities) and to generatively fill unmeasured parts of the space using an informed model (i.e. interpolate and extrapolate in a physics-aware manner). 
Considerable work would be required to recast existing methods to handle the dimensionality and different constraints of scientific data in physical parameter spaces; but the efficient representations being developed for simulating the real world may well hold useful insights for representing other kinds of coherent data-spaces.

\subsubsection{Literature Discovery}

The scale of the research literature is continually growing, making it increasingly difficult for researchers to maintain awareness of important trends or singular results. 
Conversely, this enormous scientific corpus is a trove of insights that should be more fully leveraged. 
Literature Based Discovery (LBD)\cite{Smalheiser2012LBD, HENRY201720, Thilakaratne2019LBD} has a long history of increasingly sophisticated methods and software being developed for mining the literature, identifying connections across domains, and otherwise streamlining literature research. 
AI methods, and LLMs in particular, are well-positioned to greatly accelerate these processes, automating knowledge extraction from publications.\cite{Krenn2022} 

An obvious use-case is to systematically search through a corpus in order to extract and tabulate values for quantities of interest.\cite{Young2018datamining, D4DD00051J, chiang2024llamp} Here, the flexibility of LLMs can enable extraction that handles the heterogeneity arising from synonyms, different definitions or units of measure, and so on. 

LLMs can be combined with document retrieval (RAG) to allow users to rapidly identify relevant documents (or sub-sections thereof) and immediately incorporate them into reasoning or question-answering. RAG LLMs have been used to build domain-specific chatbots for science,\cite{Yager2024chatbot} and to provide an interface to vast materials data that can be distilled as requested by the user.\cite{chiang2024llamp} 
More generally, the AI model can be exploited as a co-pilot to help the user exploit specialized knowledge or tools, as has been demonstrated for catalyst research,\cite{ramos2023bayesian} chemistry experiments,\cite{Boiko2023Nature} and chemistry tools.\cite{bran2023chemcrow} 
A valid concern is that the reasoning of LLMs---being well below human scientists---would be insufficient to be useful. However, when designed as a co-pilot, such systems can offer substantial value. 
LLMs can exploit the systematic compositionality of language (and thus ideas), which enables them to generalize in useful ways.\cite{Lake2023} 
Evidence shows that dialoging with LLMs can indeed help researchers.\cite{liang2023large, ai4science2023impact}

LLMs can also be exploited to automate tedious tasks. For instance, they can be used for ranking,\cite{qin2023large, Yager2024chatbot, evans2024largelanguagemodelsevaluators} evaluating,\cite{fu2023gptscoreevaluatedesire} or classifying\cite{Yager2024chatbot} scientific documents. This opens up a new possibility for researcher engagement with the literature, beyond the conventional activities of periodically searching for articles of interest and keeping a watch for relevant articles through networks (peers or automated). 
LLMs could be used to search, organize, rank, triage, and summarize papers, and thereby identify the most pertinent publications for human consideration. 

LBD has a strong history of exploiting network analysis to understand the science corpus, including using predictive knowledge networks.\cite{InfraNodus, Krenn2023} An interesting possibility would be to exploit modern foundation models as another form of network analysis. The semantic embedding provided by these models could offer a rich means of identifying connections (or lack thereof) in the literature. For instance, clusters of publications that are semantically similar but not cross-citing one other could represent inefficiency (duplicative efforts unaware of each other), while clusters that are highly correlated in a subset of embedding dimensions (but divergent in others) could represent opportunities for collaboration.

Another use for AI agents is to aid researchers in drafting scientific manuscripts. LLMs are fundamentally text-generation systems, and their role in productively generating long-form textual content is being extensively studied.\cite{yang-etal-2022-re3, Fitria2023ArtificialI, Pavlik2023} As often observed with LLMs, the quality of output can be improved through iteration, including using the LLM to generate an outline, self-critique output, and so on. There are additional challenges in using LLMs to generate scientific text, as consistency and correctness must not be compromised. Here too, there has been progress in using LLMs to automatically generate full-length technical documents.\cite{ALTMAE20233, lu2024aiscientistfullyautomated, shao2024assistingwritingwikipedialikearticles} 
The use of AI to generate text for inclusion in the scientific literature could be deleterious to science if the texts contain too many errors (compared to the human baseline).
The exocortex design emphasizes the central role of the human researcher in assessing correctness and validating decisions/generations. We propose that the human researcher maintain the important role of validation, and thereby maintain responsibility for the quality of publications to which they attach their name.

\subsubsection{Autonomous Ideation}

\begin{figure*}
\centering
  \includegraphics[width=5.0cm]{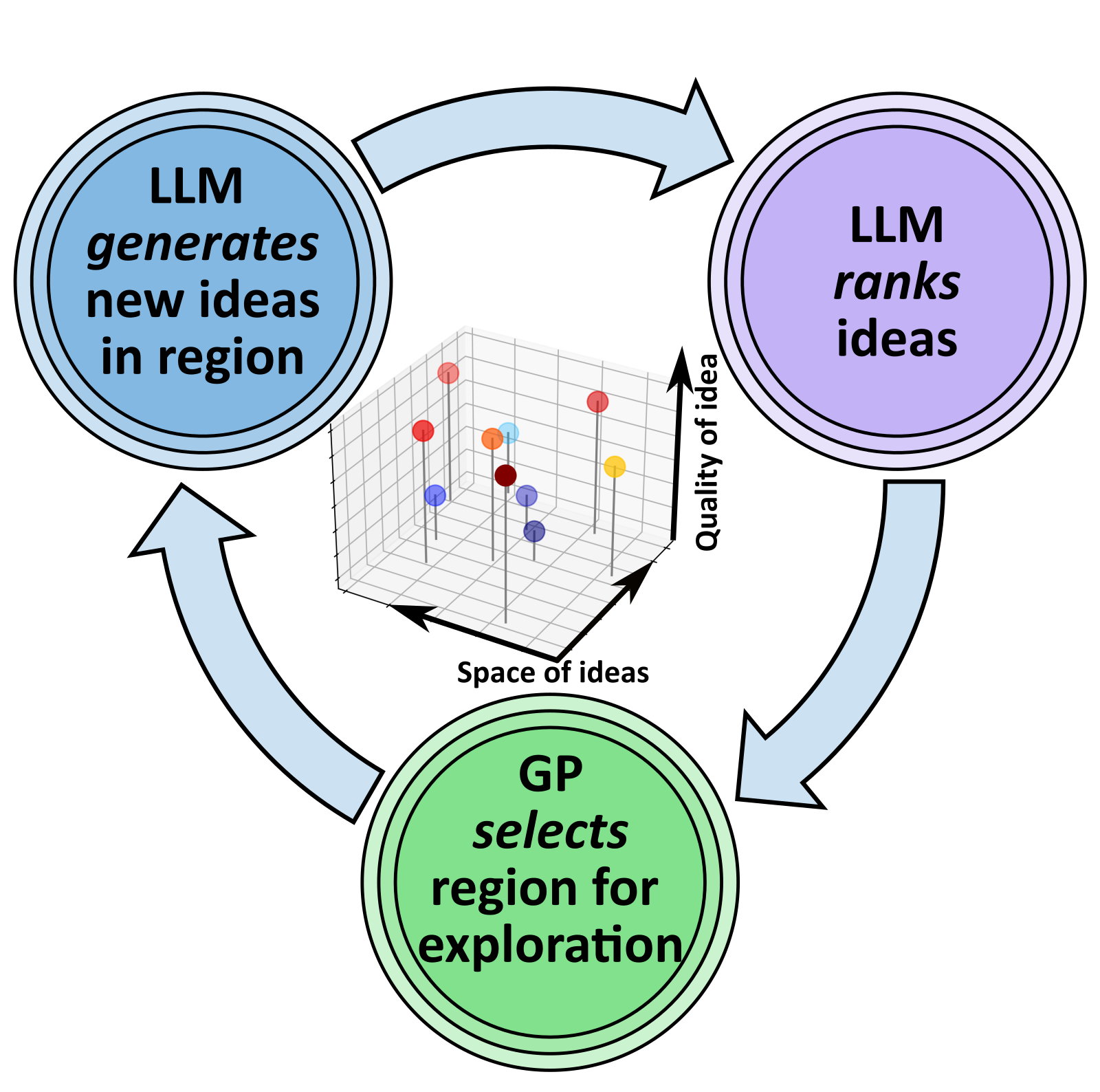}
  \caption{Autonomous ideation aims for the AI agent to develop new scientific ideas (novel research directions, testable hypotheses, actionable research plans). One possible system design is to treat the task simialr to an autonomous experimentation loop, wherein one is exploring a multi-dimensional parameter space. In ideation, one can define the space of ideas using embedding vectors to position each idea. Each idea can be scored using an LLM ranking procedure. The loop consists of selecting a region for exploration (e.g. based on some combination of local sparsity, model error, and quality-maximization), generating ideas in that region (e.g. using an LLM provided with documents/ideas from the local neighborhood), and ranking the resultant ideas. As the loop proceeds, the space becomes populated with ideas. The top generations can eventually be presented to the human for consideration.
}
  \label{fgr:ideation}
\end{figure*}

A novel use for LLMs would be to help automate the task of generating and evaluating scientific ideas, including research plans, testable hypotheses, experimental plans, and predictive theories. These cognitive tasks are among the most high-level performed by human scientists, and as such least likely to be fully automated by LLMs in the foreseeable future. On the other hand, the process of human ideation involves many secondary cognitive activities that could be automated.\cite{kambhampati2024llms} Thus, autonomous ideation seeks to generation loops of machine-driven brainstorming and evaluation, bringing high-value ideas to the human's attention for further consideration.

Existing work in LBD has begun to tackle the question of how to use natural language processing and LLMs for hypothesis generation or other scientific ideation tasks. 
A central question is whether LLMs can be creative at all. LLMs are trained statistically on a large document corpus, and can be viewed as generating novel text that are interpolations in a semantic space. Such generations can be factual (correctly composing ideas in the training data) or erroneous ``hallucinations'' (or confabulations). 
Hallucinations can be partially mitigated by detecting them through generation uncertainty,\cite{Farquhar2024} or by grounding responses using RAG.\cite{2020RAG,Yager2024chatbot,gao2024retrievalaugmented,yu2024evaluation} 
Although hallucinations are generally undesirable, their existence is intrinsic\cite{kalai2024calibrated} and there is a tradeoff between hallucinations and creativity.\cite{mohammadi2024creativity} In other words, some amount of hallucination is desirable, to enhance creativity and communication.\cite{sui2024confabulation} 
More broadly, evaluations of LLM creativity suggest that they can generate outputs that are non-trivially novel and useful to humans.\cite{Koivisto2023, haase2023artificial, girotra2023ideas, boussioux2023crowdless, doshi2023generative}
Language models have demonstrated utility for hypothesis generation,\cite{manning2024automated, ma2024dreureka} or as generators for novel ideas.\cite{wang2023learning, wang2024scimon, kambhampati2024llms} 

The most direct way to use LLMs for ideation is as a chatbot assistant to a human researcher. A more automated design would leverage agentic AI operating in loops, so that a group of LLMs propose and critique ideas, and then rank\cite{qin2023large, Yager2024chatbot} these ideas in order to identify the most promising. 
A more structured (but speculative) approach is to treat the task of autonomous ideation as being analogous to autonomous experimentation,\cite{Yager2023AutonomousChapter} wherein an ML decision-making algorithm selects points in a physical parameter space for measurement. 
In autonomous ideation, one could analogously select points in the semantic ``space of ideas'' for exploration (Figure~\ref{fgr:ideation}). More specifically, the search space is defined using a semantic vector (e.g. text embedding) and the target signal in that space is defined using LLM ranking of the ideas. On each loop, a new region is selected for exploration, using a modeling process that can consider both idea ranking (bias towards high-quality regions), and uncertainty (explore under-sampled or high-error regions). This modeling can exploit Gaussian process methods to naturally capture uncertainty and learn hyper-parameters that describe the semantics being explored. Once a point is selected, an LLM generates new ideas at that position by (e.g.) sampling a local neighborhood\cite{wang2024scimon} of ideas or documents in order to generate new content. This generation is ranked to quantify it as a signal, which is fed back into the loop.
As this procedure continues, it will naturally fill the semantic space of ideas, balancing between exploration and exploitation, and providing a surrogate model for idea quality in the subspace selected for search. 
It is an open question whether Bayesian modeling can meaningfully be applied to the inherently vague space and signals associated with ideation. But the AE framework provides a robust starting point for rigorously testing various idea exploration schemes.

A different ideation design would be to leverage ongoing work in visualizing and interpreting the internal state of the LLM. 
While neural networks are often described as inscrutable black boxes, there has been enormous progress in interpreting their structure and the latent spaces in which they operate. In vision models, the role of neurons and circuits can be interpreted by visualizing strong activation patterns.\cite{olah2017feature} 
In language models, tasks learned in-context can be understood as a simple function vector that capture the relevant input-output behavior.\cite{hendel2023incontext, todd2024function} 
A particular direction in the model's internal state can be associated with specific behavior, such as refusal to respond\cite{Arditi2024Refusal} (allowing that behavior to be selectively amplified or weakened). 
Identifying internal circuits associated with particular concepts allows one to build ``circuit breakers'' to suppress undesired output.\cite{zou2024improving} 
Natural hierarchies of concepts---which occur throughout natural language and especially in scientific ontologies---are represented in the model's internal vectorial space as polytopes that can be decomposed into simplexes of mutually-exclusive categories.\cite{park2023linear, park2024geometry} 
Model activations can be interpreted using human concepts, if they are projected into a higher-dimensional space to disentangle them.\cite{bricken2023monosemanticity, templeton2024scaling, gao2024scaling}
These interpretability insights are often exploited for alignment,\cite{wang2023aligning} to elicit safe and desirable model behavior. However, they could also be used to directly explore the landscape of ideas.
For instance, visualizing the internal ontology for a scientific sub-space might allow researchers to identify regions of unexplored concepts, or to see fruitful cross-connections between ideas that are typically considered unrelated. Searching the structure of this space for common patterns could further reveal new connections or universal motifs. Being able to directly alter the activation or geometry of the semantic space, and observing LLM output, provides another avenue for generating novel ideas in a highly directed way. 
This research thrust would require researchers building new intuitions about how to understand and navigate the complex spaces internal to LLMs.

More generally, strategies originally intended for model tuning or alignment could all be co-opted for ideation. For instance, one could block off exploration of ideas known to be fruitless, or conversely emphasize desired modes-of-thought. Viable strategies include fine-tuning,\cite{hu2021lora, dettmers2023qlora, liu2024dora} RLHF\cite{ziegler2020finetuning, ouyang2022training, lambert2022illustrating} (including AI-assisted\cite{lee2023rlaif}), constitutional adherence,\cite{bai2022constitutional} preference ranking,\cite{song2024preference}, instruction backtranslation,\cite{li2024selfalignment} principle-driven self-alignment,\cite{sun2023principledriven} or eliciting latent knowledge.\cite{pfau2023eliciting, belrose2023leace, belrose2023eliciting}

\subsection{Exocortex System}

The proposed exocortex design will behave as a system of interconnected AI agents, some of which can also communicate directly with the human researcher. The correct design for this system is an open research and engineering challenge. 
Nevertheless, we can begin to propose and test designs. One of the simplest implementations would be for each researcher to build a personalized network by selecting among pre-existing AI agents, and defining connections between them based on desired workflows. Communication between agents could thus be managed with point-to-point message queues. This approach is not very scalable, however. An alternative would be to establish a central database where inter-agent messages are accumulated, and build code that manages communications, using user-defined heuristics to decide when incoming messages require returning to the same agent for revision, launching a new agent, passing to a running agent, or bringing to the human's attention. 
Likely there are yet better designs possible if one treats the agent-interaction problem as a large machine-learning task. By selecting a flexible design (e.g. based on graph neural networks), an automated optimization process could create/eliminate connections in order to build dynamic workflows. Much of the work on within-agent iteration and looping can be exploited to improve inter-agent workflows.

In all these schemes, signals between agents can take the form of plaintext messages. This has the advantage of being highly legible to the human operators,\cite{Aschenbrenner2024} allowing them to understand commands, make improvements, and even extract scientific value from intermediate products. 
As the number of agent types increases, the diversity of possible inter-agent cooperations increases quadratically, while the space of possible workflows grows exponentially. 
Example messages that might be sent between agents are shown in Table~\ref{tab:messaging}. 
Legible inter-agent messages will allow the human operator to inspect, at will, operation of the system, including editing an agent's message before it is executed by another agent.

The complexity of interconnected agents, and the non-standardized (text-based) messaging between them poses a problem for automated monitoring, analysis, and optimization of these systems. On the other hand, it is possible that existing approaches for systems engineering can be recast to the context of AI swarms. For instance, machine-learning has benefiting enormously from gradient backpropagation,\cite{Rumelhart1986} which has essentially automated the process of optimizing complex neural network and AI models. By analogy, Yuksekgonul et al. proposed TextGrad as a text-based ``differentiation'' of AI systems.\cite{yuksekgonul2024textgrad} Natural language feedback (e.g. criticism) of system outputs can be used as scores (analogous to loss), the variation in score as a function of changes in prompt can be used as a gradient, and gradients can be propagated across the system with knowledge of architecture. This allows automated optimization of LLM-interaction networks. 
Zhou et al. demonstrate how symbolic learning can be applied to optimizing LLM frameworks.\cite{zhou2024symboliclearningenablesselfevolving} 
Further developing techniques such as these may be crucial to properly optimizing exocortex-like systems.

\begin{table*}[ht]
    \centering
    {\small
    \begin{tabularx}{\textwidth}{p{0.12cm} p{1.5cm} M{0.7cm} |L L L L L L}
        & & & \multicolumn{6}{c}{\textbf{Message from}} \\
        & & & Autonomous Experimentation & Experimental Assistant & Data Exploration & Knowledge Mapping & Literature Discovery & Autonomous Ideation \\
        & & &  \makecell[c]{\includegraphics[width=0.5cm]{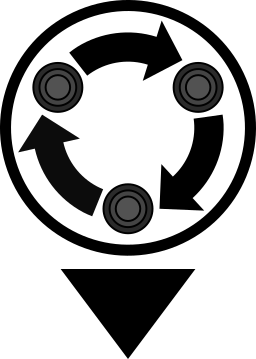}} & \makecell[c]{\includegraphics[width=0.5cm]{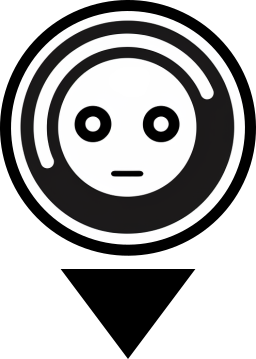}} & \makecell[c]{\includegraphics[width=0.5cm]{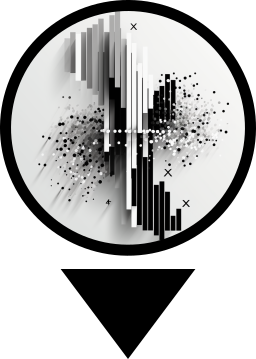}} & \makecell[c]{\includegraphics[width=0.5cm]{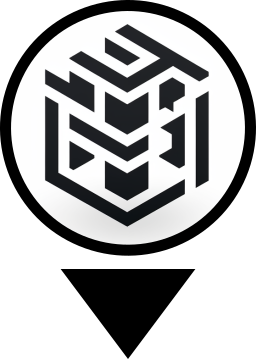}} & \makecell[c]{\includegraphics[width=0.5cm]{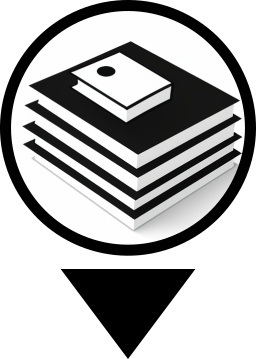}} & \makecell[c]{\includegraphics[width=0.5cm]{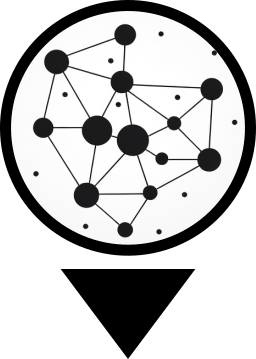}} \\ \hline
        %
        & Autonomous Experimentation 
        & \includegraphics[height=0.5cm]{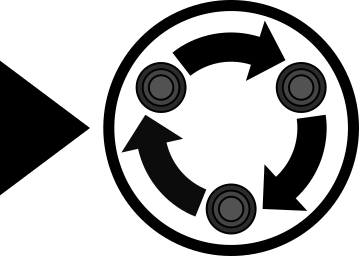}
        & \textcolor{kgrey}{Relaunch this AE with tweaked parameters...}
        & Launch new AE with these parameters...
        & Incorporate this data as AE prior...
        & Build AE kernel that conforms to this model...
        & Set AE parameters based on this literature...
        & Launch AE that tests this idea... \\ \cline{2-9}
        %
        & Experimental Assistant 
        & \includegraphics[height=0.5cm]{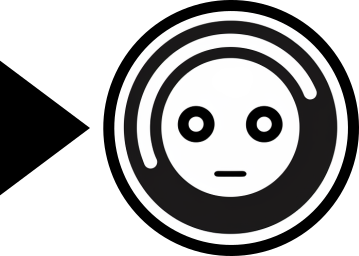}
        & Queue these follow-up experiments...
        & \textcolor{kgrey}{Launch this sub-experiment...}
        & Analyze these related datasets...
        & Overlay this model with current experiment... 
        & Create plan that replicates this published experiment...
        & Prepare an experiment plan to test this idea... \\ \cline{2-9}
        %
        \multirow{4}{*}{\rotatebox{90}{\textbf{Message to}}} 
        & Data Exploration 
        & \includegraphics[height=0.5cm]{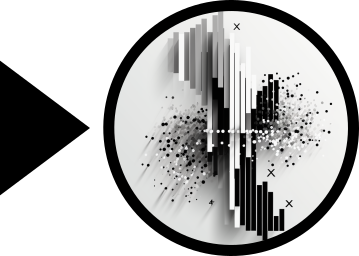} 
        & Plot ongoing AE...
        & Retrieve and analyze data for this experiment...
        & \textcolor{kgrey}{Compare with this dataset...}
        & Overlay this model on dataset...
        & Add citations to dataset...
        & Add annotations to dataset... \\ \cline{2-9}
        %
        & Knowledge Mapping & \includegraphics[height=0.5cm]{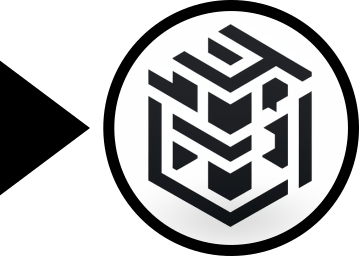} 
        & Test integration of AE into model...
        & Add this experiment to model...
        & Integrate this data into model...
        & \textcolor{kgrey}{Compare this model with past models...}
        & Integrate literature results into model...
        & Annotate model with this interpretation... \\ \cline{2-9}
        %
        & Literature Discovery & \includegraphics[height=0.5cm]{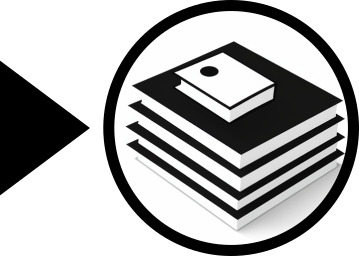} 
        & Is this trend expected, based on prior work...
        & Search literature for the value of...
        & Find data values relevant to...
        & Find models/equations relevant to...
        & \textcolor{kgrey}{Find and summarize papers on this sub-topic...}
        & Retrieve literature relevant to this topic-space... \\ \cline{2-9}
        %
        & Autonomous Ideation & \includegraphics[height=0.5cm]{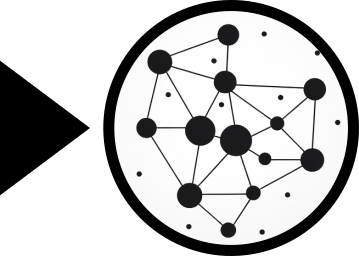} 
        & Evaluate progress of this ongoing AE...
        & Generate hypotheses relevant to current experiment...
        & Generate hypotheses for this dataset...
        & Propose theories to explain this model...
        & Launch ideation based on these papers...
        & \textcolor{kgrey}{Launch ideation on this sub-topic...} \\ \hline
    \end{tabularx}
    }
    \caption{Examples of command messages that various AI agents could send to other agents. The diagonal elements (grey text) are commands sent from an agent to another instance of the same type.}
    \label{tab:messaging}
\end{table*}

The proposed exocortex architecture would treat agents as modules, allowing them to be swapped or for new agents to be added. Connecting agents to each other should enable progressively more complex automated workflows. However, a crucial open question is whether multi-agent workflows can scale to complex problems. For instance, even with a low per-step error rate, long task sequences could easily accumulate intolerable total error rates. 
The acceptable error-rate will be quite different for different parts of a workflow. For instance, imperfect ideas generated during ideation have low risk, as they will be identified and filtered out by the human easily (in human ideation there is value in initially considering erroneous ideas, as this can improve creativity). On the other hand, errors introduced by a data-analysis agent could be subtle and difficult to detect; yet errors in this stage would contaminate downstream analysis and thus invalidate the science. Errors in the experimental stage could waste valuable resources (time, experimental material, etc.) but are likely to be caught by human oversight. 
AI agent workflows can also be difficult to debug (owing to stochastic response) and brittle to maintain (changes in cloud models, changes in input data distribution, etc.). 
Thus, the scalability of multi-agent workflows is a crucial open question, requiring research and development. 
Frontier work in this area suggests that well-designed multi-step AI workflows may be able to generate coherent outputs.\cite{lu2024aiscientistfullyautomated}


The goal of the exocortex is to augment a human scientist's intelligence. 
This objective is predicated on the assumption of emergence at two levels: one, that the swarm of AI agents will, through coordination, exhibit intelligence greater than the naive summation of their respective abilities; and two, that the combination of exocortex agents and human thinking will enable greater effective intelligence. 
To succeed, the exocortex architecture must thus enable this outcome. The correct design remains an open research question. However, we propose that analogies to human cognition can aid in the design. 

\subsubsection{AI-AI Interactions}

LLMs generate ideas and decisions, but they are quite primitive in the sense that the ideas are reflexive rather than resulting from deep introspection.\cite{Aschenbrenner2024} The repeated waves of processing that occur within an LLM as it proceeds through tokens provides an opportunity to build-up more complex assessments, with the current understanding represented as updates to the residual stream. Improved behavior can thus be elicited by inducing the model to explicitly output reasoning steps.\cite{wei2023chainofthought, kojima2023large} 
Interestingly, introducing even meaningless filler tokens into the output provides improved performance,\cite{pfau2024lets} presumably owing to the additional computation cycles that are invoked. And yet, LLMs implement a relatively primitive and unidirectional method of thinking, as they are unable to revise the serialized output. 
Multiple research efforts aim to improve this by introducing a sort of deliberation cycle, such as by triggering self-critique of output,\cite{shinn2023reflexion, lightman2023lets} or generating chains of thought through iterative self-prompting.\cite{xu2023reprompting, yao2023tree, xu2024faithful} 
Exploiting tree search (e.g. Monte Carlo) can further improve quality, especially on math problems.\cite{luo2024improve, chen2024alphamath, zhang2024restmcts, zhang2024accessing}
For scientific applications, versions of these methods that explicitly invoke formal logic are especially attractive.\cite{xu2024faithful, vashishtha2024teachingtransformerscausalreasoning} One can also provide pre-designed thought-templates to improve reasoning on selected tasks;\cite{yang2024buffer} building a catalog of templates for scientific tasks would be beneficial.

Another means of generating improved output is to construct ``societies'' of semi-specialized AI agents, and allow them to communicate and cooperate on a task. The hope is that specialization improves diversity and allows task-specific targeting, and that the emergent quality of collective output is higher than for any individual agent. 
Although this approach is only nascent,\cite{li2023camel, chen2023agentverse, hong2023metagpt, park2023generative, zhuge2023mindstorms, frisch2024llm, guo2024large, wang2024mixtureofagents} there are early suggestions that it can improve task performance in contrived contexts (e.g. games\cite{wang2023describe, abdelnabi2023llmdeliberation}) and applications (e.g. code generation\cite{dong2024selfcollaboration} and translation\cite{wu2024perhaps}). 
One can also use synthetic analogs of cultural transmission to improve learning of AI swarms.\cite{Bhoopchand2023, perez2024cultural} 
Interaction between agents can be\cite{guo2024large} cooperative, debating, or competitive. Agents can be organized into flat structures, where each agent is equivalent (e.g. voting on answers/decisions), or hierarchically, where top-level agents assign tasks to workers, and aggregate outputs. Different tasks will, of course, call for different organizational structures. However, there are often clear advantages to establishing hierarchies and workflows,\cite{zhuge2023mindstorms} especially where one can draw inspiration from human organizational structures. 

Instead of emulating human social structures, an alternate architecture is mixture-of-agents,\cite{wang2024mixtureofagents} which organizes AI blocks into layers reminiscent of neural networks (where each node is an LLM instead of a synapse). The input prompt is fed into a layer of models that propose independent responses, an aggregator synthesizes the responses into an improved output, and this is  is fed into the next layer. Thus, response quality progressively improves across layers, as more reconsideration is performed. By including different LLMs within a layer, one can improve diversity and allow for models to compensate for each other's weaknesses. Performance can also be optimized by correct selection of models within layers. The architecture is rationalized and organized, and amenable to rescaling (changing number of agents per layer, number of layers, etc.) to optimize for a particular task. 
This work demonstrated significantly improved outputs, compared to single-shot use of any underlying model, and demonstrated that a final aggregation LLM call (rather than ranking and selecting the best output so far) improves generation. 
This supports the idea that agent interactions can lead to emergent capabilities greater than any individual agent. The iterative processing may also make multi-agent setups amenable to longer-horizon tasks (e.g. longer text analysis or generation). 

The optimal architecture for providing LLMs with deliberative capabilities remains an open and exciting research question.\cite{Aschenbrenner2024} Current scaling suggests that LLMs have untapped potential that could be unlocked with appropriate designs. 
In parallel with algorithmic research, we propose that scientific researchers can make progress by simply expending compute to compensate for architectural weaknesses. For instance, consider tasking an agent-swarm with a problem, whereupon the agents generate ideas, ask each other questions, generate random permutations by combining ideas, rank all ideas, and only present the best ideas to the human. This workflow is highly wasteful in the sense that the majority of the generated content is never seen and indeed low-quality. Yet this invisible content can be viewed as the system's internal deliberation. Even if this process is inefficient, its automated and unobtrusive nature can make the outputs sought-after by humans. In the context of science expenditures, the associated costs could be small relative to the value. 
There are tentative reports that this kind of extended search can yield substantial improvements.\cite{jones2021scaling, agarwal2024manyshot, Greenblatt2024ARCAGI, hassid2024largerbetterimprovedllm, brown2024largelanguagemonkeysscaling, snell2024scalingllmtesttimecompute, wu2024empiricalanalysiscomputeoptimalinference}
We thus propose increasing investigation by physical scientists of such brute-force workflows, for generating content useful to researchers.

\subsubsection{Human-AI Interactions}
Human thinking involves a combination of effortless intuition and deliberative reasoning\cite{Kahneman1982, Stanovich_West_2000, Kahneman2003, EVANS2003454} (often referred to as ``implicit'' vs. ``explicit'' or as ``system 1'' vs. ``system 2''). 
A cluster of low-level brain modules generate reflexive actions, intuitive assessments, and creative ideas. A higher-level deliberative process engages in discrimination, iterative refinement, and selection; using the low-level generators as inputs and assessors. 
A synthetic exocortex can be designed similarly. The swarm of AI agents act as low-level generators, introducing ideas and providing reflexive assessments. The human deliberative consciousness remains the core, doing the highest-level discrimination and decision-making, and is thus ultimately the locus of volition. 

The exocortex interface should ideally make the AI-generated inputs feel much like the human's own low-level modules. When actively working on a task, the exocortex should provide contextual assessments and ideas that feel like spontaneous intuition that the human can trust (but will also verify). 
When returning to a dormant task, accumulated background AI-swarm processing should feel like the mental incubation known to occur in humans,\cite{Sio2009} wherein returning to a problem after a diversion often yields new insights and perspective owing to subconscious consideration.

Obviously, efficient coupling between reflexive and deliberative processes is required in humans for effective creativity and problem-solving.\cite{Beaty2015} 
A legitimate concern is that traditional peripheral-based user interfaces (using keyboards, screens, etc.) represents too much friction for strong coupling, and brain-computer interfaces will be required.\cite{6893912} 
However, there is ample evidence of human tool use becoming overlearned\cite{Driskell1992Overlearning} to the point that the tool is considered an extension of the person's body and volition.\cite{Maravita2004Tools} We can view the evolution of cognitive technology as precedent for humans externalizing aspects of their cognition, with a succession of tools (writing, calculators, the Internet, smartphones) being exploited as external memories, processing extensions, or task-activation schemes. 
Thus, we posit that fast and responsive interaction through existing computer interfaces may be sufficient for the desired interaction. Indeed, humans are known to be able to enter so-called ``flow states'' (immersed and focused)\cite{Csikszentmihalyi1990, Ellis1994Flow} during computer-oriented tasks such as programming.\cite{Gold2020Flow}

\subsubsection{Human-Computer Interface}

The purpose of the exocortex is to offer the human additional cognitive power that feels---as much as possible---as a natural extension to their own mind. One can imagine a future where brain-computer interfaces are used to provide and ideal interface;\cite{6893912} we posit that in the short term much value can be realized by providing researchers with AI agents through traditional computer interfaces. Research in human use of autonomous tools suggest that the person must ultimately feel that they are in control of processes.\cite{AutonomySurveyOpinions} 
Correspondingly, we propose that initial exocortex interfaces will involve humans reviewing and verifying LLM plans before they are executed (by other AI agents). 
There is evidence that humans observing the output of LLMs debating each other helps the human identify the best ideas.\cite{michael2023debate, khan2024debating} This suggests that, more generally, providing researchers with access to exocortex inter-communications (critique, debate, refinement, etc.) could provide them with valuable information. 
As system robustness improves, and user confidence in the tools increases, more and more workflows can be automated and unattended. 

With respect to human interaction with the software tools, we can define several modalities:
\begin{itemize}
  \item \textbf{Push}: Where alerts are used to capture the user's attention (operating system notifications, text messages, etc.).
  \item \textbf{Pull}: Which require the user to actively check on status (visiting web page, opening a program, etc.).
  \item \textbf{Ambient}: Where information is displayed peripherally to the user, or where contextually relevant.
\end{itemize}

Different aspects of exocortex operation might imply a different notification mode. For instance, human-directed dialogue is inherently \textit{pull}, while operationally-critical and time-sensitive statuses that require human resolution will be \textit{push}. However, the \textit{ambient} modality is the most well-aligned to the ethos of the exocortex, where information generated by AI agents is contextually but unobtrusively presented, available to subconscious consideration by the human, and thus appears to the user as a seamless extension of their ongoing planning.

In the short term, we can envision useful interfaces being developed by exploiting HCI best-practices for ambient information display, and by integrating exocortex outputs into existing visualization tools and workflows. 
Extended reality (virtual reality, augmented reality) tools may be natural peripherals for exocortex software. 
Leveraging improving systems for voice transcription and voice synthesis provides another avenue for natural interaction with these tools. 
We note that as LLMs increase in capability, they are beginning to develop a primitive theory of mind.\cite{kosinski2023theory, Webb2023, Strachan2024mind, street2024llms}
This can be taken advantage of by using the LLM to roughly model human behavior, and thereby providing suggestions in ways that are most beneficial and least disruptive.

\subsection{Infrastructure}
In addition to novel AI developments, the success of the exocortex requires continued progress in several pragmatic infrastructure components (left side of Figure~\ref{fgr:exocortex}). 
In general, science infrastructure must be made increasingly automated and software-accessible, so that AI agents will be able to leverage these systems as tools. 
Importantly, even if the exocortex concept is flawed, the proposed improvements in science infrastructure will be of great value to the community.

\subsubsection{Automated Instruments}
As previously discussed, scientific instruments are becoming increasingly automated. 
This trend is driven by the increasing complexity of these tools (there are too many layers of control for them all to be manually managed), and researcher desires for speed and efficiency. 
Automated tools are in principle amenable for activation by AI agents. The primary limiting factor is the availability of an external API for both triggering actions (synthesis or measurement), and retrieving results (raw or analyzed data). 
We encourage researchers and tool vendors to push aggressively towards a world wherein every piece of laboratory equipment has an API, and is thus amenable for AI automation. 
LLM technology may in fact be a crucial enabler for such a transition, since their ability to handle arbitrary and heterogeneous APIs (as long as documentation is provided) liberates researchers and manufacturers from having to agree on and follow a single standard for laboratory automation.

Conversely, it must be acknowledged that automation of scientific instruments and laboratory workflows represents a bottleneck for AI-driven science. While AI models and software can be rapidly iterated and improved, hardware system improvements are more capital-intensive and require longer-timescale design and construction efforts. 
Vendor-provided tools may use proprietary data formats and may not expose software interfaces that provide complete control of the system. These represent significant roadblocks to automation; the community should correspondingly demand commercial solutions that adhere to open data standards. 
Although mechanizing and automating laboratory work is by no means trivial, we argue that the value of any such effort will increase dramatically in the coming years, as AI agent control systems increase in sophistication.

\subsubsection{Open Science Databases}
There is growing appreciation that the data underlying scientific publications should be open and freely available to others. 
Open data practices increase the realized value of a research effort, as datasets can be used by others in ways not originally envisioned.\cite{Connolly2021National, Yager2023AutonomousChapter} 
For example, datasets can be used for meta-analysis, to identify broader trends, and as inputs to machine-learning training. 
The FAIR data principle emphasizes that all datasets should be Findable, Accessible, Interoperable, and Reusable.\cite{Wilkinson2016FAIR} 
In practice, this means data must be retained and archived, that archives should be open for download and indexing, and that data should be correctly labelled and have corresponding meta-data to contextualize and associate it (with people, groups, publications, and related datasets). 

The exocortex is closely tied to open data efforts. To function most effectively, it requires that AI agents be able to identify and operate on vast datasets. Thus, the exocortex is empowered by the greater availability of research data of all types. Obviously, the exocortex also improves as more domain-specific AI modules are trained; this will typically require aggregating openly available domain datasets. 

The exocortex concept can also potentially improve data release. One key limiter in data release is researchers being unable to provide sufficiently detailed meta-data, because common tools lack meta-data features and because of the time burden associated with manually adding human annotations to vast datasets. AI agents can help here, as they are better able to handle the ambiguity of sparse meta-data labelling. AI agents may also be able to help automate the collection of meta-data and annotation of datasets when they are first produced, which should increase the richness of meta-data captured. The lesson for the community to learn is that data release is valuable even if the dataset is imperfectly organized and annotated. Future tools will make it possible to organize and extract value even from heterogeneous and unlabelled data.

\subsubsection{Software}
Software underlies an enormous amount of scientific research, and its importance is further growing as more machine-learning methods are integrated into science. 
LLMs can play an important role in scientific software, for code generation\cite{D3DD00113J} and code execution by calling APIs\cite{yao2023react, schick2023toolformer, gao2023pal, liang2023taskmatrixai, shen2023hugginggpt, cai2023large, peng2023check, xu2023rewoo, hsieh2023tool} or interacting with graphic user interfaces (GUIs).\cite{lai2024autowebglm, pan2024autonomous, xie2024osworld, bai2024digirl}
LLMs could also play a role in user education, since they provide a way for scientists to learn new software systems via chatbot assistance or LLM generation of code exemplars.

Greater integration of scientific software tools into AI agent workflows will require these tools to be made readily available. Fortunately, the prevailing trend in scientific software is to release code as open source, and make it available via repositories; these make it possible for automated systems such as AI agents to take advantage of them. AI agents may well be able to handle the some of the heterogeneity of modern software deployments; that is, they may be able to automatically download code, set up an appropriate containerized environment, generate wrapper code for interacting with that container, and then activate the system. However, it may well be preferable for the community to begin developing and adopting flexible but standardized methods of containerizing scientific software so that it can be more easily shared and launched. For instance, the MLExchange effort is developing a web platform for working with containerized ML models.\cite{MLExchange}

We also note a substantial software infrastructure challenge. Running a large number of AI agent instances, and enabling coordination between them, is a substantial technical challenge. Integrating these resources into existing scientific software workflows is also challenging.\cite{FosterReview} The expertise of the high-performance computing (HPC) community can be leveraged, as existing know-how with respecting to scaling and deployment of large-scale science software systems should be transferable to AI swarm architectures.

\subsubsection{Publications}
As with data, publications should ideally be broadly open in order for an exocortex to leverage them. 
It must be possible for the exocortex to identify relevant documents, retrieve them, and read them. Currently, the vast majority of the scientific literature is not easily available for machine indexing, retrieval, and AI/ML training. 
Researchers will generally not be able to negotiate the required licenses with publishers in order to easily obtain access to the required. 
Luckily, scientific practices have been increasingly moving towards open access, where the publication (or at least a preprint version of it) is freely available. 
We encourage researchers and publishers to continue pushing in this direction, since it both benefits traditional scientific practices, and helps enable future AI-driven workflows.

In addition to policy questions, there are engineering tasks required in order to make the scientific literature readily available to AI agents. In the short term, researchers may need to manage these activities themselves, building a curated local corpus of documents for their agents to interact with. In the medium term, AI agents will likely be able to access publications through tools or APIs. In the long term, the community would ideally build a common AI database on top of the existing literature. For instance, a community database that kept track of embeddings for every published paper (and sub-sections thereof) would avoid the wasted cost of researchers recomputing embeddings when their own agent ingests publications. 
Such a database could also store AI-generated secondary products associated with papers (summaries, classifications, connections to other literature, proposed research directions, etc.). Sharing such a database would allow each researcher's exocortex to leverage the work of all other researchers exocortices.

\subsubsection{Facilities}
\begin{figure*}
\centering
  \includegraphics[width=10.0cm]{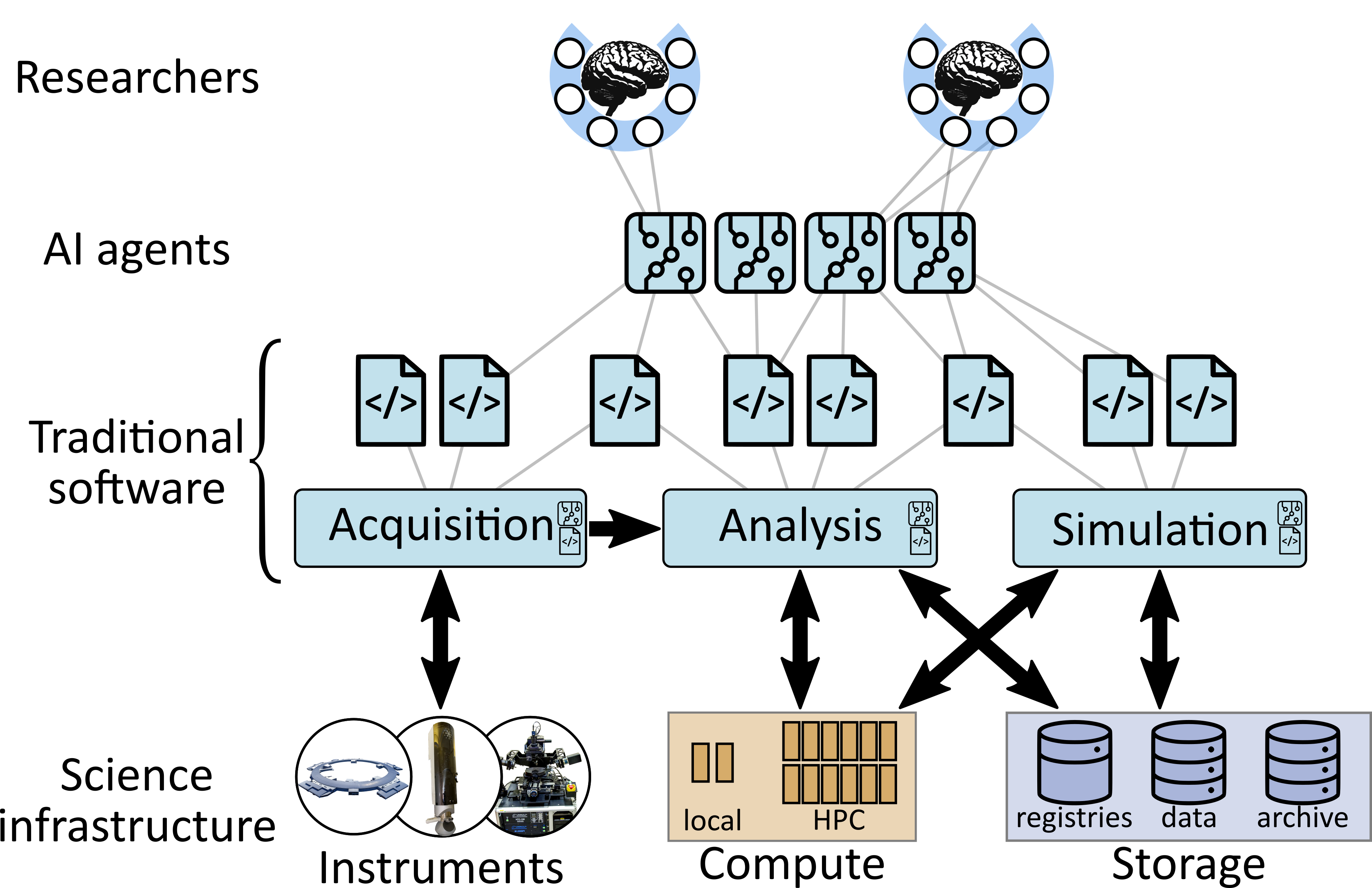}
  \caption{A possible architecture for AI agents aggregating access to scientific facilities. Each researcher's exocortex could negotiate control by dialoging with a set of AI agents provided by the facilities. That layer of agents would be optimized to launch tasks using traditional software APIs. The underlying resources (measurement instruments, compute resources, databases) would be triggered and queried, with the outputs integrated first by the facility AI agents, and then by the researcher's exocortex.
}
  \label{fgr:facilities}
\end{figure*}

Scientific research tools are often organized into coherent facilities that offer multiple related capabilities, or multiple versions of a particular measurement tool (as in the case of electron microscopy centers, synchrotrons, FELs, etc.). 
As more synthesis, processing, and measurement tools become individually automated and collectively organized into facilities, we can begin to imagine the impact that agentic AI will have on them. 
In particular, agentic AI will enable a transition of scientific facilities away from individual tools that are selected and micro-managed by scientists, and into a discovery ecosystem, wherein users can phrase their high-level scientific goals, and rely upon a swarm of AI agents to correctly select tools, launch experiments, and aggregate results. 
A possible architecture is depicted in Figure~\ref{fgr:facilities}. Each researcher's exocortex, which knows about that researcher's scientific goals and problem-specific science constraints, can negotiate with AI agents operated by the facilities. This allows researchers to conceive of science goals, and leverage agents to convert this into actionable plans. 
The scientific facilities design and operate AI agents responsible with providing access to a variety of systems, and correctly coordinating between these systems. For instance, a particular research goal might require launching a set of measurement tools and corresponding simulations, and then aggregating the results to compare. This coordination could be executed by a a combination of AI agents and traditional software infrastructure. 
This vision inherently requires ubiquitous and reliable automation of individual systems. It also requires novel developments in research infrastructure, to more efficiently cross-connect between components. 
We postulate that agentic AI will be an enabling technology for accomplishing this interconnection, as it will bypass the need for every sub-component to adhere to a single standard for meta-data and communication. As long as each component provides a documented software interface, the layer of AI agents should be able to productively access it. 
The productivity gain from this architecture could be transformative, as it would allow researchers to conduct experiments of a complexity previously impossible.

\section{Perspectives}

We have presented an admittedly speculative vision for the future of science, wherein each scientist has a personalized exocortex---a swarm of AI agents working together to automate research and expand researcher cognition. 
While this vision currently seems far-fetched, it is now within reach owing to recent developments in LLMs; and it becomes increasingly realistic as LLM technology improves. 
Indeed, the exocortex is envisioned in such a way that it automatically leverages improvements in the technology of AI agents, as more powerful models can be swapped in progressively as they become available. 
We propose that the science community should work together, and aggressively pursue the creation of systems like this.

We suggest that physical scientists focus on applications of AI agents, and learning how best to connect agents into coherent workflows. In fact, science is an ideal proving ground for agentic AI, since scientists can articulate precise goals, assess rigor of reasoning, and evaluate success. Thus, AI/ML researchers will hopefully view the physical sciences as an ideal environment in which to research agents and agent swarms.

The proposed multi-agent interactions and workflows highlight several open research questions. It is not known whether complex multi-step AI tasks will be sufficiently robust. The community must measure how AI capabilities scale, as a function of task complexity and inter-agent organizational architecture. 
The bottlenecks for scientific discovery---especially automated discovery workflows---must be elucidated. We speculate that LLMs will provide high utility for ideation and hypothesis generation, by providing the human with text digests and ranked ideas, and by acting as a conversational partner. However, integration with experimental tools is likely to lag, owing to the time and cost associated with building and testing laboratory automation systems. With respect to the overall exocortex system, we envision the largest roadblocks arising from managing the complexity of inter-communicating agents, and establishing sufficient reliability.

We emphasize that the proposed work is valuable even if the exocortex concept turns out not to be the right framing. 
The proposed improvements to science infrastructure---making it increasingly robust, automated, software-accessible, and auditable---has value even if AI agents are not successful. 
The proposed AI agents---streamlining access to publications, data, software, and instruments---are valuable even if their interconnection into an exocortex proves fruitless.

The science exocortex has enormous potential impact. 
There is growing evidence that generative AI methods exhibit various forms of emergence, including world modeling,\cite{li2023emergent, akyürek2023learning, kosinski2023theory, Webb2023} concept generalization,\cite{Ganguli_2022, wei2022emergent, nanda2023progress, bubeck2023sparks, Webb2023} and pattern aggregation that is more capable than the inputs.\cite{zhang2024transcendence} 
The exocortex architecture would enable and leverage additional layers of emergence. Interactions between AI agents should lead to more reliable, coherent, and capable output than single-shot generation by a lone LLM. And, crucially, interaction between a swarm of AI agents---each responsible for intelligently mediating access to a suite of research capabilities---and a human researcher should lead to the emergence of enhanced human capabilities. By expanding the researcher's intelligence into the exocortex, the researcher can accomplish more, as they are able to intuitively and seamlessly weave myriad physical, computational, and cognititve systems into their intellectual work.

\section*{Author Contributions}
KGY reviewed the literature, developed the concepts, and wrote the manuscript.

\section*{Conflicts of interest}
There are no conflicts to declare.

\section*{Acknowledgements}
This research was carried out by the Center for Functional Nanomaterials, which is a U.S. DOE Office of Science Facility, at Brookhaven National Laboratory under Contract No. DE-SC0012704. We thank Dr. Charles T. Black for fruitful discussions.


\bibliographystyle{rsc}
\bibliography{rsc}

\end{document}